%% file: main.tex
\documentclass{article}

\usepackage[a4paper, total={6in, 8in}]{geometry}

\usepackage{subcaption}
\usepackage{float}
\usepackage{multirow}
\usepackage{amsmath}

\usepackage{hyperref}
\usepackage{authblk}
\usepackage{comment,url}

\usepackage{subcaption}
\usepackage{makeidx}         % allows index generation
\usepackage{graphicx}        % standard LaTeX graphics tool
\usepackage{multicol}        % used for the two-column index
\usepackage[bottom]{footmisc}% places footnotes at page bottom

\usepackage{newtxtext}       % 
\usepackage[varvw]{newtxmath}       % selects Times Roman as basic font

\usepackage{float}
\usepackage{multirow}

\title{Discovering Partially Known Ordinary Differential Equations: a Case Study on the Chemical Kinetics of Cellulose Degradation}
\author[1]{Federica Bragone}
\author[2]{Kateryna Morozovska}
\author[3]{Tor Laneryd}
\author[4]{Khemraj Shukla}
\author[1]{Stefano Markidis}
\date{January 2025}

\affil[1]{KTH Royal Institute of Technology, Lindstedtsv\"gen 5, 100-44, Stockholm, Sweden}
\affil[2]{KTH Royal Institute of Technology, Malvinas v\"ag 10, 100-44, Stockholm, Sweden}
\affil[3]{Hitachi Energy Research, V\"aster\aa s, Sweden}
\affil[4]{Brown University, 170 Hope Street, Providence, USA}

\begin{document}

%\and Matthieu Barreau \at KTH \email{}
%\and Kateryna Morozovska \at KTH, Address of Institute \email{kmor@kth.se}}
\maketitle
%star abstract is the online abstract
\begin{abstract}
    The degree of polymerization (DP) is one of the methods for estimating the aging of the polymer based insulation systems, such as cellulose insulation in power components. The main degradation mechanisms in polymers are hydrolysis, pyrolysis, and oxidation. These mechanisms combined cause a reduction of the DP. However, the data availability for these types of problems is usually scarce. This study analyzes insulation aging using cellulose degradation data from power transformers. The aging problem for the cellulose immersed in mineral oil inside power transformers is modeled with ordinary differential equations (ODEs). We recover the governing equations of the degradation system using Physics-Informed Neural Networks (PINNs) and symbolic regression. We apply PINNs to discover the Arrhenius equation's unknown parameters in the Ekenstam ODE describing cellulose contamination content and the material aging process related to temperature for synthetic data and real DP values. A modification of the Ekenstam ODE is given by Emsley's system of ODEs, where the rate constant expressed by the Arrhenius equation decreases in time with the new formulation. We use PINNs and symbolic regression to recover the functional form of one of the ODEs of the system and to identify an unknown parameter.
\end{abstract}

\input{Introduction}

\input{Method}
\input{Results}

\bibliographystyle{plain}
\bibliography{Reference} 

\end{document}

%% file: Introduction.tex
\section{Introduction}
\label{sec:introduction}
Modeling the degradation of materials is challenging since aging and degradation are complex processes influenced by many factors. It is hard to collect data for prolonged aging processes, such as the aging of electric insulation, as it takes part over a long period of time (10-60 years) and can be impacted by the environment and the operation of the equipment itself. The studies on the degradation and aging of electric insulation are often limited, and collected measurements require more explainability, which we can now gain by using novel machine learning methods.  

This work will further analyze the mechanisms of degradation and aging in electric insulation, using an example of thermally upgraded paper or cellulose, which is immersed in mineral oil and is used as an insulation system in power transformers. 
Several deterioration processes affect cellulose aging, including oxidation, hydrolysis, and pyrolysis~\cite{cigre2007ageing}. These processes often depend on each other, and it is difficult to isolate each degradation mechanism and study its impact directly. Therefore, degradation of electrical insulation is usually measured using a reference parameter named the Degree of Polymerization (DP)~\cite{fabre1960deteriorating, shroff1985review, emsley1992reassessment}, which represents the number of monomeric units in the polymer and decreases over time, because the polymer chains break with higher exposure to aging mechanisms. Therefore, DP values and the paper condition are highly correlated; the initial value of DP ranges between 1200 and 1000 for new equipment~\cite{cigre2007ageing}, and it decreases continuously over time. A DP value of approximately 200~\cite{fabre1960deteriorating, shroff1985review} is typically considered end-of-life due to paper aging for the insulation and the electrical component. 

The scarcity of cellulose aging data challenges the accurate tracking and prediction of paper conditions. 
Therefore, finding alternative models that can work with a limited quantity of data and extrapolate the required information to simulate different scenarios is necessary. Physics-Informed Neural Networks (PINNs)~\cite{raissi2019physics} can overcome the problem of limited data. They are Artificial Neural Networks (ANNs) that embed the physical equations in their architecture. While classical ANNs are heavily dependent on data, PINNs exploit the physical laws described by Ordinary and Partial Differential Equations (ODEs and PDEs) and can show good performance even for small and noisy datasets. Apart from their ability to solve ODEs and PDEs using forward problem formulation, 
PINNs can solve the inverse problem by finding the parameters that best describe the given data or, in other words, discover the parameters of given ODEs and PDEs~\cite{karniadakis2021physics}.

The aging of cellulose in electric insulation systems is usually modeled using ODEs. 
The Ekenstam equation gives one model to estimate the DP degradation inside power transformers~\cite{ekenstam1936behaviour}. It is based on the statistical models for cellulose degradation developed by Freudenberg and Kuhn et al.~\cite{freudenberg1930hydrolyse}. In this ODE, the reaction kinetic constant is modeled according to the Arrhenius equation, which relates the reaction rate to the absolute temperature~\cite{emsley1992reassessment}. It introduces parameters like the activation energy $E$ and the pre-exponential factor $A$ that change for different materials. A modification of the Ekenstam kinetic model is later introduced by Emsley et al.~\cite{emsley1997kinetics}, where the reaction rate is no longer considered constant. Emsley considers a system of ODEs where the reaction rate constant has its differential equation and decreases using first-order kinetics. Several equations were introduced afterward to improve the accuracy of the degradation models~\cite{zervos2005cotton, ding2008degradation, calvini2008kinetics}. More accurate analysis and comparison of the different models can be found in~\cite{calvini2014meaning, mendez2024kinetic}. However, more studies are needed to identify a better system of equations. 

Modeling the degradation of cellulose insulation requires solving differential equations with partially known parameters.
Our previous work focused on estimating one unknown parameter of the Arrhenius equation, using PINNs with synthetic data~\cite{bragone2022physics}. In the current study, we propose some techniques to help discover the equations that accurately model the DP degradation. We use PINNs to find better parameters and functions given observed field data. Using PINNs ability to discover equation parameters, we infer unknown parameters of the Arrhenius equation in the Ekenstam kinetic model, particularly the activation energy $E$ and the pre-exponential factor $A$.
Later, we combine PINNs with symbolic regression to discover an unknown function of Emsley's system of ODEs and simultaneously estimate the unknown parameter. In this case, the PINN model uses an extra neural network to approximate the values of the unknown function while inferring the unknown parameter. Finally, we rediscover the unknown function using symbolic regression. This idea follows from work by Zhang et al.~\cite{zhang2024discovering}, where authors apply PINNs in combination with symbolic regression to generic dynamical systems and also to find a reaction-diffusion model for a real-world application. Symbolic regression is a powerful tool that explores the space of feasible mathematical expressions and finds the best fit that describes the data, respecting accuracy and simplicity~\cite{billard2002symbolic}.

The contributions of this paper are the following:
\begin{itemize}
    \item We investigate techniques to discover partially known ODEs to estimate the cellulose degradation in power transformers. 
    \item We estimate the unknown parameters of the Arrhenius equation implemented in the Ekenstam equation testing both with synthetic data and real DP data points.
    \item By assuming we do not know the form of the ODE modeling DP in Emsley equations, we estimate both the function and the rate constant using first PINNs and then symbolic regression to retrieve the form of the unknown function. 
    \item We list these techniques' possible advantages and drawbacks to discover differential equations for aging power components.
\end{itemize}

%% file: Method.tex
\section{Methodology}
\label{sec:methods}
This section describes in greater detail two cellulose degradation mechanisms commonly used for modeling aging. Later, we introduce the PINNs used in this study and explain the procedure for discovering unknown parameters and a function. 

\subsection{Degradation Models of Power Transformers}
\label{sec:background}
We introduce the equations that model the aging of power transformers, which we consider in this work.
The Ekenstam equation is expressed as:
\begin{equation}
\begin{aligned}
    & \frac{d\text{DP}}{dt} = -k\cdot \text{DP}^2, \\
    & \text{DP}_0 = \text{DP}(0),
    \label{eq:ekenstam_ode}
\end{aligned}
\end{equation}
where DP$_0$ is the DP at time 0, the initial condition, and $k$ is a constant. The equation states that the number of unbroken chain bonds available is proportional to the reaction rate at any time (right-hand side).
A relationship between the slope of the reaction kinetic constant $k$ from Eq. \eqref{eq:ekenstam_ode} and the temperature is established with the Arrhenius equation~\cite{emsley1992reassessment,emsley1994kinetics}. The Arrhenius equation is defined as follows:
\begin{equation}
    k = Ae^{-\frac{E}{RT}},
    \label{eq:arrhenius_eq}
\end{equation}
where $A$ is the pre-exponential factor and determines the cellulose contamination from moisture, water, acids, and oxygen; $E$ is the activation energy, which indicates the temperature dependence of the cellulose aging process and is expressed in [J/mol]; $R$ is the gas constant [8.314 J$\text{K}^{-1}\text{mol}^{-1}$]; $T$ is the absolute temperature in Kelvin [K]. 
The activation energy $E$ does not depend directly on the reaction conditions. For example, the value of the pre-exponential factor increases drastically as the degradation conditions deteriorate in the presence of water and oxygen.

A modified version of the kinetic model delineated by Ekenstam was then introduced by Emsley et al.~\cite{emsley1997kinetics}. In their model, the reaction rate $k$ is no longer a constant but decreases as aging occurs. The second degradation model, i.e., Emsley's system of ODEs, is given by:
\begin{equation}
\begin{aligned}
    & \frac{d\text{DP}}{dt} = -k_1\cdot \text{DP}^2 , & \quad  & \text{DP}_0 = \text{DP}(0),\\
    & \frac{dk_1}{dt} = -k_2\cdot k_1 , & \quad & k_{1_0} = k_1(0),
    \label{eq:emsley_syst}
\end{aligned}
\end{equation}
where $t$ is the time expressed in hours, $k_1$ is the reaction rate, $k_2$ is the rate constant at which $k_1$ deteriorates, and DP$_0$ and $k_{1_0}$ are the DP and $k_1$ values at time 0, i.e. their initial conditions. 

\subsection{Discovery of Unknown Parameters}
\label{subsec:disc_unknown_par}
 We solve the inverse problem for the Ekenstam ODE described in Eq.~\eqref{eq:ekenstam_ode}. Therefore, given some observed DP values spanning a transformer's lifetime, we would like to infer the two parameters in the rate constant expressed by the Arrhenius equation, precisely the pre-exponential factor $A$ and the activation energy $E$. Two parts define the PINNs model. The first is an NN that takes time coordinates as inputs corresponding to the observed data points in the dataset. The output approximates the DP values at the inputted time coordinates. The second part consists of the evaluation of the ODE, where the output of the NN is used to calculate the derivatives with respect to time using automatic differentiation for the residual function $f$. Therefore, the residual function is defined in the following way:
\begin{equation}
    f = \frac{d\text{DP}}{dt} + Ae^{-\frac{E}{RT}}\text{DP}^2.
\end{equation}
To train the model, we define the loss function as the mean-squared error MSE:
\begin{equation}
    \text{MSE} = \text{MSE}_{data} + \text{MSE}_{ode},
    \label{eq:loss}
\end{equation}
where
\begin{align}
    & \text{MSE}_{data} = \frac{1}{N}\sum_{i=1}^{N}{|\hat{\text{DP}}(t_{\text{DP}}^i)-\text{DP}^i|^2}, \quad \label{eq:MSE_data} \\
    \quad & \text{MSE}_{ode} = \frac{1}{N}\sum_{i=1}^{N}{|f( t_{\text{DP}}^i)|^2}. \label{eq:MSE_ode}
\end{align}
MSE$_{data}$ corresponds to the loss function assigned to the observed data points, and MSE$_{ode}$ is the loss function for the residual. $\{t_{\text{DP}}^i,\text{DP}^i\}_{i=1}^N$ is the set of training data points on $\text{DP}(t)$, $\hat{\text{DP}}$ represents the PINNs approximation for DP, and $N$ is the number of training data. The number and position of the residual $f$ collocation points are the same as the training data points. 

The values for the parameters $A$ and $E$ are large. Parameter $A$ ranges between values of order $10^8$ and $10^9$, while for $E$ is around $10^5$~\cite{emsleystevens1994kinetics, lundgaard2004aging, cigre2007ageing, lelekakis2012ageing1, lelekakis2012ageing2}.  
Given the complexity of the equation's values, it is necessary to scale the equation to guarantee a proper convergence for the PINNs model. If the parameter values are too large, PINNs have a wider region to consider and struggle to converge to the global minimum. For this reason, we introduce a scaling of the equation as follows:

\begin{equation}
    f = \frac{d(ln(\text{DP}))}{dt} + ln(A) - \frac{E}{RT} + 2\cdot ln(\text{DP}).
    \label{eq:ln_dp_equation}
\end{equation}

\begin{figure}[ht]
    \centering
    \includegraphics[width=1.0\textwidth]{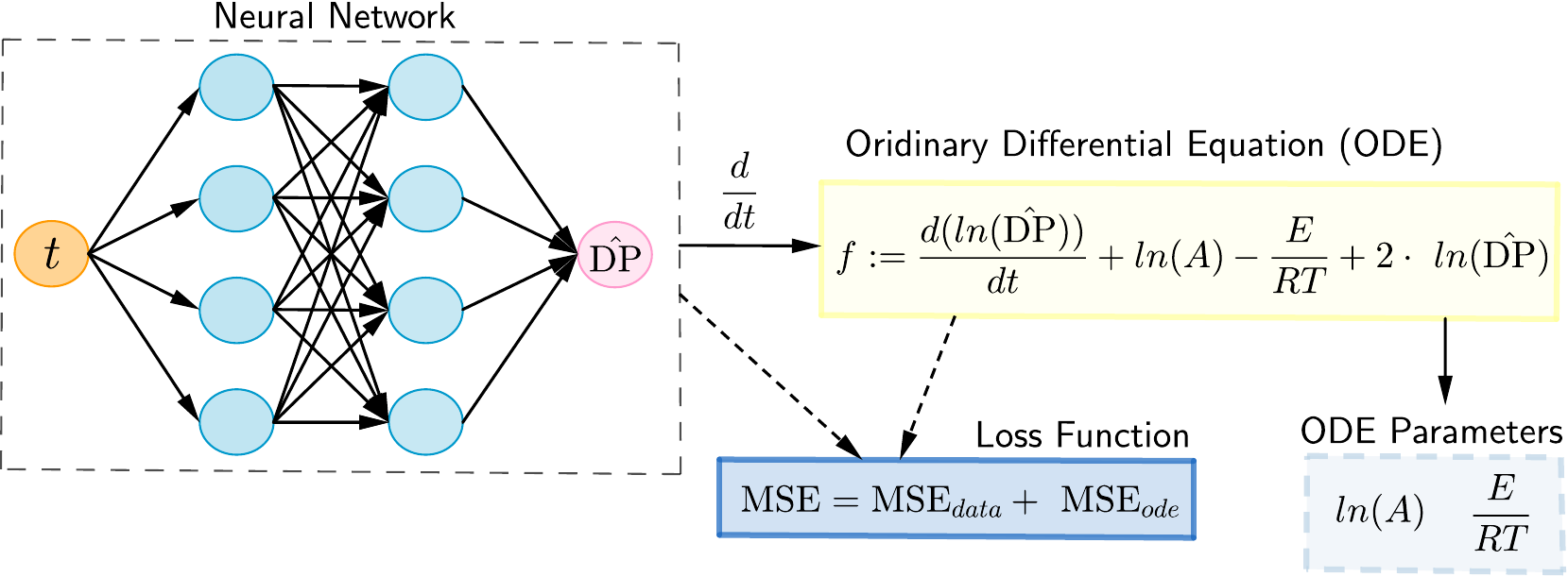}
    \caption{Schematic representation of PINNs for the discovery of the scaled unknown parameters $ln(A)$ and $\frac{E}{RT}$ of the Ekenstam ODE. }
    \label{fig:pinns_schema_unknown_params}
\end{figure}

Using the natural logarithm $ln$, we can bring almost all the parameters to reasonable values that can be considered. However, using the natural logarithm, we still have the value $E$ to deal with in the equation. Therefore, we look at the term $\frac{E}{RT}$ instead to discover with the PINNs model. The parameters that we aim to infer from Eq.~\eqref{eq:ln_dp_equation} are $ln(A)$ and $\frac{E}{RT}$, which we will refer to as our scaled parameters. Fig.~\ref{fig:pinns_schema_unknown_params} shows the structure of PINNs.

We first test our model on two cases with synthetic data and then a case with DP measurements. For the synthetic data, we solve the ODE using the integration package from the SciPy Python library~\cite{virtanen2020scipy} with $A=3.42\cdot 10^8$, $E=1.1\cdot 10^5$ J/mol, $T=352$ K, $R=8.314$ J$\text{K}^{-1}\text{mol}^{-1}$ and the initial value of DP, $\text{DP}_0=1100$. The transformer reaches the 200 DP threshold in almost 30 years with these values. We simulate two datasets of 24 and 48 equally distant points and consider a period of 40 years. Moreover, we add 1\%, 5\%, and 10\% of Gaussian noise in the datasets to test the models' performance and the predictions' robustness. The data is also scaled since the first part of the PINN model consists of an NN dealing with the data points. In particular, the time $t$ was scaled between values 0 and 1, while the DP values were divided by a factor of 100. This scaling was also considered in the ODE, especially for the time, as the equation deals with time in hours, and the data is reported in years. Therefore, the ODE was also multiplied by a factor of 100 and 24(hours) $\times$ 365(days) $\times$ 40(years), as the last term is the final year considered. Finally, the natural logarithm of these scaling parameters was taken.

The measurements dataset consists of 22 DP data points obtained from~\cite{pradhan2005estimation}. In the paper, the authors retrieved DP values along with other properties through several long-duration aging experiments and reported the mean values of the measurements. As the dataset starts around the first year, an initial value of DP of 1100 for time $t=0$ is assigned to indicate new transformer paper, which is usually the case. The values for the pre-exponential factor $A$ and the activation energy $E$ are not reported for this dataset. The literature shows specific range values where the parameters $A$ and $E$ lie~\cite{emsleystevens1994kinetics, lundgaard2004aging, cigre2007ageing, lelekakis2012ageing1, lelekakis2012ageing2}. We consider the parameter $A$ ranging in values of the order $10^8$ and $10^9$, while for the parameter $E$, we take 111$\pm$6 kJ/mol. Therefore, knowing these ranges, we can understand if the results obtained align with what is known in the field. We scaled the data similarly to the synthetic data cases, considering that the final year of the reported data is 35.27 years.

\subsection{Discovery of an Unknown Function and a Parameter}
\label{subsec:disc_unknown_func_par}
\begin{figure}[ht]
    \centering
    \includegraphics[width=0.8\textwidth]{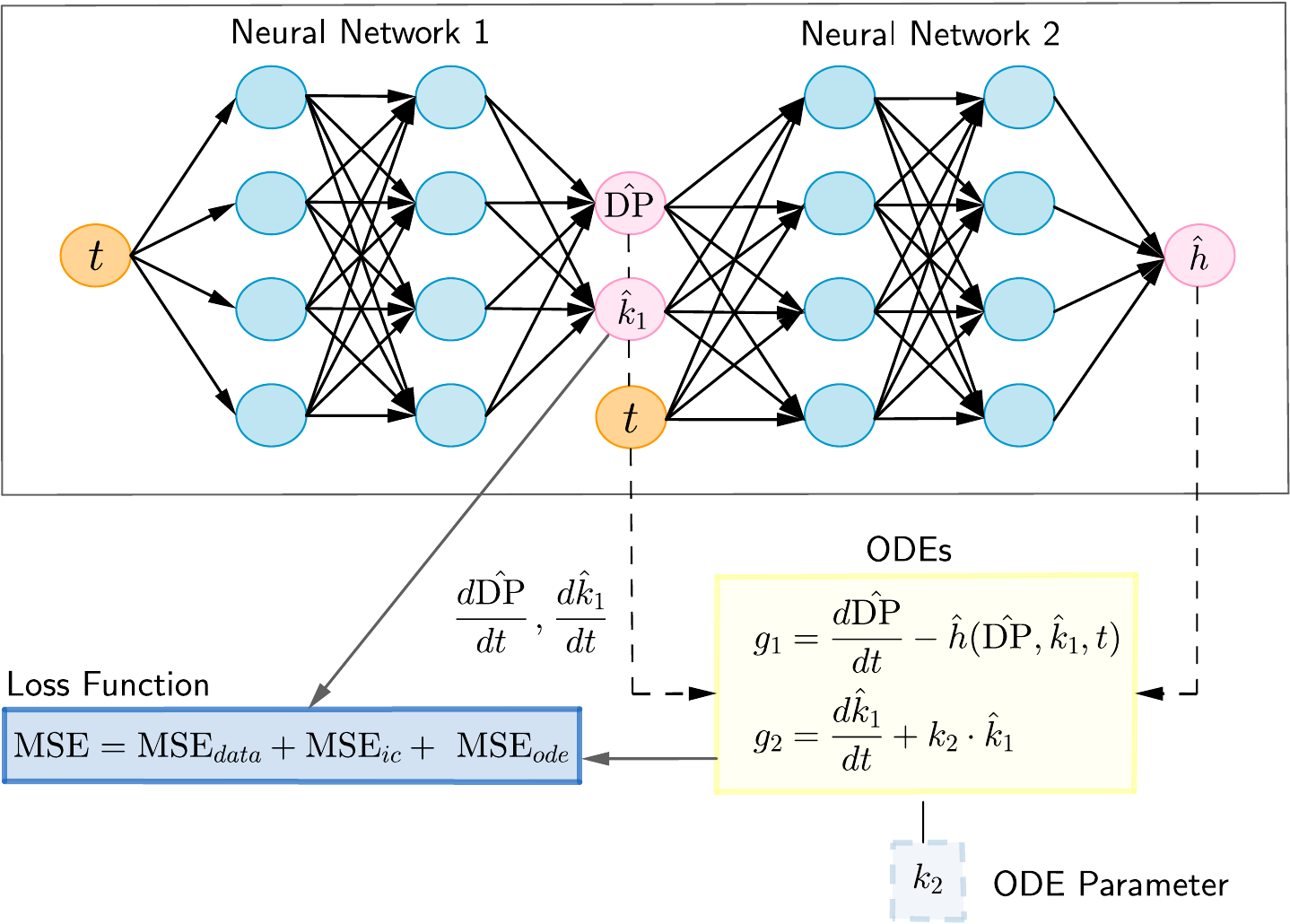}
    \caption{Schematic representation of PINNs for the discovery of the unknown function $h(\text{DP},k_1,t)$ and parameter $k_2$ of the Emsley system of ODEs.}
    \label{fig:pinns_schema_unknwons_paramsfunct}
\end{figure}

As described in Sect. \ref{subsec:disc_unknown_par}, we can infer unknown parameters of equations given observed data. However, it is often the case that we do not know the exact form of the differential equation that we are considering. Given some observations, it is also possible to recover the functional form of the equation. In our case, we assume we only know one part of Emsley's system of ODEs expressed in Eqs.~\eqref{eq:emsley_syst}. In particular, we assume that the differential equation modeling the DP values is unknown.
Moreover, we also assume the value of the parameter $k_2$ is unknown. Again, we define a PINN model to solve the inverse problem. However, in this case, the PINN has a slightly different architecture, illustrated in Fig. \ref{fig:pinns_schema_unknwons_paramsfunct}. The first part, which consists of the NN approximating the solution, is extended here with an additional NN to approximate the unknown function. After, the approximations of the two NNs are incorporated into the ODE evaluation part of the PINN. The two NNs have their own sets of weights and biases. The training happens at the same time as the PINN training, and the form of the function can be learned from the data. We define the residual functions as follows:
\begin{equation}
\begin{aligned}
    & g_1 = \frac{d\text{DP}}{dt} - h(\text{DP}, k_1, t), \\
    & g_2 = \frac{dk_1}{dt} + k_2\cdot k_1,
    \label{eq:emsley_res}
\end{aligned}
\end{equation}
where $h(\text{DP}, k_1, t)$ is the unknown function that we want to estimate. 
To describe the loss function, we define $u$ and $g$ as the concatenations along the columns of the data points for DP and $k_1$ and of the functions $g_1$ and $g_2$, respectively. The loss function becomes:
\begin{equation}
    \text{MSE} = \text{MSE}_{data} + \text{MSE}_{ic} + \text{MSE}_{ode} ,
    \label{eq:loss_funcpar}
\end{equation}
where
\begin{align}
    & \text{MSE}_{data} = \frac{1}{N_u}\sum_{i=1}^{N_u}{|\hat{u}(t_u^i)-u^i|^2}, \quad \label{eq:MSE_data_funcparam} \\
    \quad & \text{MSE}_{ic} = \frac{1}{N_0}\sum_{i=1}^{N_0}{|\hat{u}(0)-u^i|^2}, \label{eq:MSE_ode_funcparam}
    \\
    \quad & \text{MSE}_{ode} = \frac{1}{N_g}\sum_{i=1}^{N_g}{|g( t_g^i)|^2}. \label{eq:MSE_ode_funcparam}
\end{align}
The construction is similar to Eq.~\eqref{eq:loss}; however, in this case, the MSE$_{data}$ includes data for DP and $k_1$ solutions, i.e. $u$, MSE$_{ode}$ includes the two residuals $g_1$ and $g_2$, i.e $g$, and we add the loss function solely for the initial conditions MSE$_{ic}$ to obtain more accuracy.
$\{t_u^i,u^i\}_{i=1}^{N_u}$ is the set of training data points on $u(t)$, which includes $\text{DP}(t)$ and $k_1(t)$, $\hat{u}$ represents the PINNs approximation for DP and $k_1$, $\{t_g^i\}_{i=1}^{N_g}$ is the set of collocation points for $g(t)$, $N_u$ is the number of training data, $N_0$ is the number of training data for the initial conditions, and $N_g$ is the number of collocation points at which the residuals are evaluated at. 

As mentioned before, the initial value of DP is around 1200-1100. The values for $k_1$ and $k_2$ are quite small. In particular, $k_1$ ranges between values of order $10^{-7}$ and $10^{-6}$, while $k_2$ is between $10^{-4}$ and $10^{-3}$. However, given the structure of the equations, we do not require scaling with the natural logarithm as for the Arrhenius equation. The scaling follows the physical units of the parameters involved. 

After obtaining an approximation of our unknown function $h(\text{DP}, k_1, t)$, we can then use symbolic regression to retrieve the exact form of the equation, as in~\cite{zhang2024discovering}. Symbolic regression is a machine learning technique that provides mathematical expressions that best describe the given data~\cite{billard2002symbolic}. This tool searches a wide space of symbolic expressions to model a relationship between the given inputs and the output. A measurement of how well the proposed equation and the data are fitted is given. This measurement is then optimized through various optimization algorithms, like genetic algorithms. Our work uses the PySR Python library~\cite{cranmer2023interpretable}. The library's optimization tool is the multi-population evolutionary algorithm that finds the mathematical expression that best fits the given 
data. The evaluation metric used to quantify the correctness of the mathematical equation is the score, defined as:
\begin{equation}
    \text{score} = -\frac{\Delta \text{log(MSE)}}{\Delta C},
\end{equation}
where MSE is the mean-squared error between the exact values and the predictions, defined as the loss for the model, $C$ is the complexity of the expression, and $\Delta$ is the local change~\cite{cranmer2020discovering}. The model should preferably have a low loss and a high score. One common way to choose the best candidate is to check the expression with the highest score between the ones with a loss lower than 2 times the best-identified expression. However, we will examine all the expressions found and choose the one with the best score with a low loss value, considering the equation's complexity.

For this part, we consider simulated data. The system of ODEs is solved using the integration package from the SciPy Python library~\cite{virtanen2020scipy}. The initial conditions considered are DP$_0=1190$ and $k_{1_0}=1.6\cdot 10^{-7}$, and $k_2=4.2\cdot 10^{-4}$ for temperature $65^{\circ}$C. These values are retrieved from~\cite{soares2001low}, where the authors reported the best-fit parameters for the Emsley equations to their observed data. They collected DP values for three different temperatures, namely $65^{\circ}$C, $80^{\circ}$C and $120^{\circ}$C, and listed the corresponding initial values of DP and $k_1$ and the constant $k_2$. The considered period is 3500 hours. We simulate 1000 training data points and use 10000 collocation points. The predictions are tested on 100 equally distant data points. We scale the data again, and in this case, the time is taken between 0 and 10, and the DP values are divided by a factor of 1000, while the $k_1$ values are scaled by $10^{-7}$. Similarly, as before, we scale the ODEs accordingly by multiplying them by these factors.
The code for this part is based on the GitHub implementation available at https://github.com/ShuaiGuo16/PINN\_symbolic\_regression

%% file: Results.tex
\section{Results}
\label{sec:results}
This section reports the results of discovering the unknown parameters of the Ekenstam ODE and discovering the unknown function and parameter of the Emsley system of ODEs. First, we show the findings for the former case. As mentioned in Sect.~\ref{subsec:disc_unknown_par}, we assess our PINN model to discover the parameters $ln(A)$ and $\frac{E}{RT}$ for two synthetic datasets consisting of 24 and 48 data points, adding also $1\%$, $5\%$, and $10\%$ of Gaussian noise, and one dataset including measurements.  
For each dataset, we perform ten model runs, changing seeds, to ensure the robustness and non-randomness of the model. 
For simplicity, we will refer to the synthetic dataset with 24 points, the one with 48 points, and the real dataset as Datasets 1,2 and 3, respectively. 
The PINN architecture consists of 3 hidden layers and 50 neurons each for synthetic data and the measurements dataset. The activation function is the sigmoid, and the network is optimized using an Adam optimizer with a learning rate of 1e-3. However, with the synthetic data, we run the model for 50000 epochs, while with the measurements dataset, we run it for 60000 epochs. The initial guesses for the scaled parameters are $ln(A)=19$ and $\frac{E}{RT}=38$. The actual values for the synthetic data are $ln(A)=19.650$ and $\frac{E}{RT}=37.587$. Therefore, the initial guesses are close to the actual values but different.
Table~\ref{tab:errors_disc_A&E} presents the relative $L_2$ errors of the solution DP, the percentage errors of the identified parameters $ln(A)$ and $\frac{E}{RT}$, and the estimated values for the scaled parameters $ln(A)$ and $\frac{E}{RT}$ and the corresponding unscaled parameters $A$ and $E$ for Dataset 1 and its three cases with added noise, Dataset 2 and Dataset 3. All the reported errors and estimations are mean values taken over the ten model runs.
The errors for Dataset 2 with added noises are omitted as they follow a similar behavior to that of the dataset with halved points. Looking at the relative $L_2$ error of the predicted DP values compared to the original data points of Dataset 1 for all four cases, including the noisy values, the values increase as more noise is added. Similar behavior can be seen for the percentage error for parameter $E$ and, in part, for parameter $A$, with an exception. The exact value for $ln(A)$ is 19.650, and the closest estimation is given by Dataset 1 with $5\%$ of noise added. 
All the models estimate the correct value of the integer part of the two inferred parameters. However, while for the estimation of $ln(A)$, most of the models get an accurate approximation of the first decimal digit, it is not the case for the parameter $\frac{E}{RT}$. The more noise is added in Dataset 1, the further the estimation of the scaled parameter for $E$ goes from the ground truth. Comparing the estimations and errors obtained with Dataset 1 and Dataset 2, there is not much difference. Therefore, a larger dataset does not necessarily involve more accurate predictions. Dataset 3 instead works with real data and does not have a reference value for the parameters $A$ and $E$. The values of the data points are in a similar range as the simulated datasets, except that the aging rate is slower, reaching the end-of-life threshold around 35 years. If we compare the estimated parameters of Dataset 3 with the ones of Dataset 1 with $10\%$ of noise, we can notice closer similarities. This could indicate that our real dataset has some noise and that our PINN model could be overfitting. 

\begin{table}
\caption{Relative $L_2$ error between the exact solution and the PINNs prediction for DP; percentage error of the identified parameters $ln(A)$ and $\frac{E}{RT}$; estimated values of the scaled parameters and their unscaled counterparts $A$ and $E$.}
\label{tab:errors_disc_A&E}
\centering
\renewcommand{\arraystretch}{1.3}

\begin{tabular}{|c|c|c|c|c|c|c|} 
 \hline 
\multirow{3}{*}{} & \multicolumn{4}{c|}{\textbf{Dataset 1}} & \textbf{Dataset 2}& \textbf{Dataset 3} \\ \cline{2-7}
 &  \multicolumn{6}{c|}{\textbf{Added Noise}} \\ \cline{2-7}
  & $\boldsymbol{0\%}$ &  $\boldsymbol{+ 1\%}$ &  $\boldsymbol{+ 5\%}$ &  $\boldsymbol{+ 10\%}$ &  $\boldsymbol{0\%}$ & $\boldsymbol{0\%}$ \\
 \hline
 \textbf{Error DP} & 8.252$\cdot 10^{-3}$ & 8.757$\cdot 10^{-3}$ & 1.464$\cdot 10^{-2}$ & 4.503$\cdot 10^{-2}$ & 9.259$\cdot 10^{-3}$ & 4.829$\cdot 10^{-3}$ \\ 
 \hline
 \textbf{Error} $\boldsymbol{ln(A)}$ & 0.139\%  & 0.166\% & 0.054\% & 0.499\% & 0.113\% & NA \\
 \hline
   \textbf{Error} $\boldsymbol{\frac{E}{RT}}$ & 0.562\%  & 0.549\% & 0.644\% & 0.867\% & 0.577\%  & NA \\
 \hline \hline
 \textbf{Est.} $\boldsymbol{ln(A)}$ & 19.623  & 19.618 & 19.653 & 19.740 & 19.628  & 19.737 \\
 \hline
 \textbf{Est.} $\boldsymbol{A}$ & 3.327$\cdot 10^8$ & 3.321$\cdot 10^8$ & 3.429$\cdot 10^8$ & 3.741$\cdot10^8$ & 3.345$\cdot 10^8$ & 3.732$\cdot 10^8$ \\
 \hline

 \textbf{Est.} $\boldsymbol{\frac{E}{RT}}$ & 37.376 & 37.381 & 37.345 & 37.261 & 37.370  & 37.265  \\
 \hline
 \textbf{Est.} $\boldsymbol{E}$ & 109381 & 109385 & 109292 & 109046 & 109365 & 109057 \\
 \hline
\end{tabular}
\end{table}

\begin{figure}[H]
    \centering
    \includegraphics[width=0.65\textwidth]{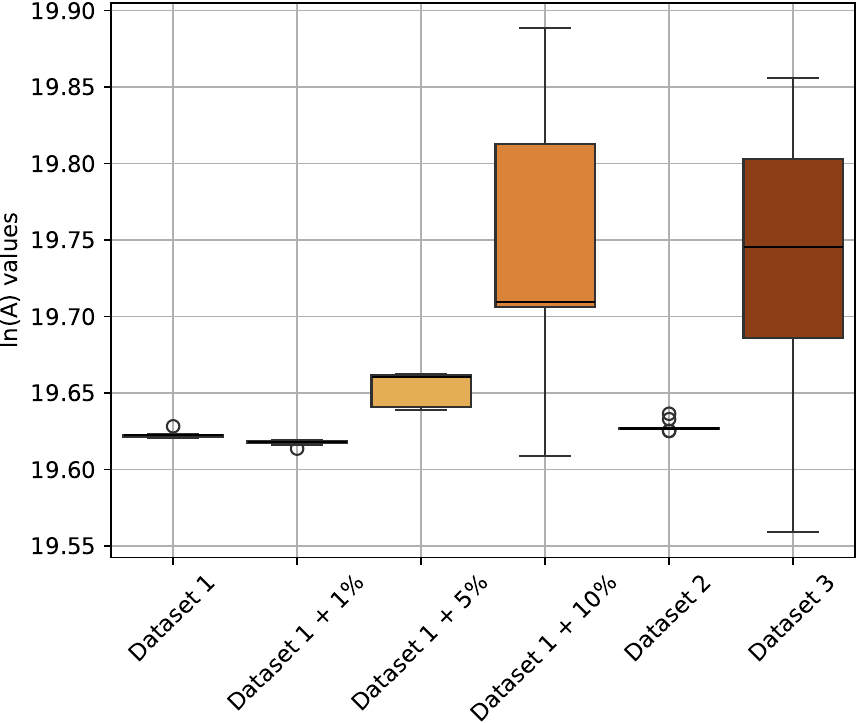}
    \caption{Box plot for the estimated $ln(A)$ values for Dataset 1, its three cases with added noise, Dataset 2 and Dataset 3.}
    \label{fig:boxplot_A}
\end{figure}

\begin{figure}[H]
    \centering
    \includegraphics[width=0.65\textwidth]{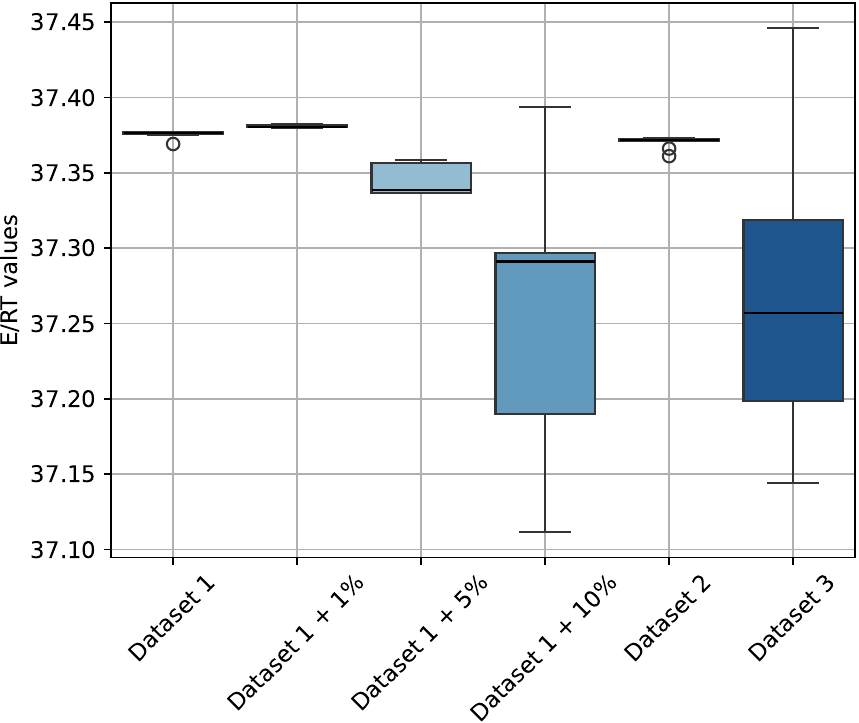}
    \caption{Box plot for the estimated $\frac{E}{RT}$ values for Dataset 1, its three cases with added noise, Dataset 2 and Dataset 3.}
    \label{fig:boxplot_E}
\end{figure}

To analyze better how robust the tested models are, we plot the box plots of the estimated scaled parameters for all the ten runs for Dataset 1 with the three noisy cases, Datasets 2 and 3. Figs. \ref{fig:boxplot_A}, and \ref{fig:boxplot_E} represent such plots for the estimation of the scaled parameters $ln(A)$, and $\frac{E}{RT}$, respectively. From both figures, we can notice that the estimations of the parameters for Dataset 1, its case with 1\% added noise, and Dataset 2 are all very similar for the different seeds except for a few outliers. 
As more noise is added, the distribution of the various seeds spreads more, indicating a larger variance. Moreover, we have a wider distribution of the values estimations for Dataset 3, probably indicating some possible presence of noise, which is consistent since it consists of measurements. 

\begin{figure}[ht]
    \centering
    \includegraphics[width=0.65\textwidth]{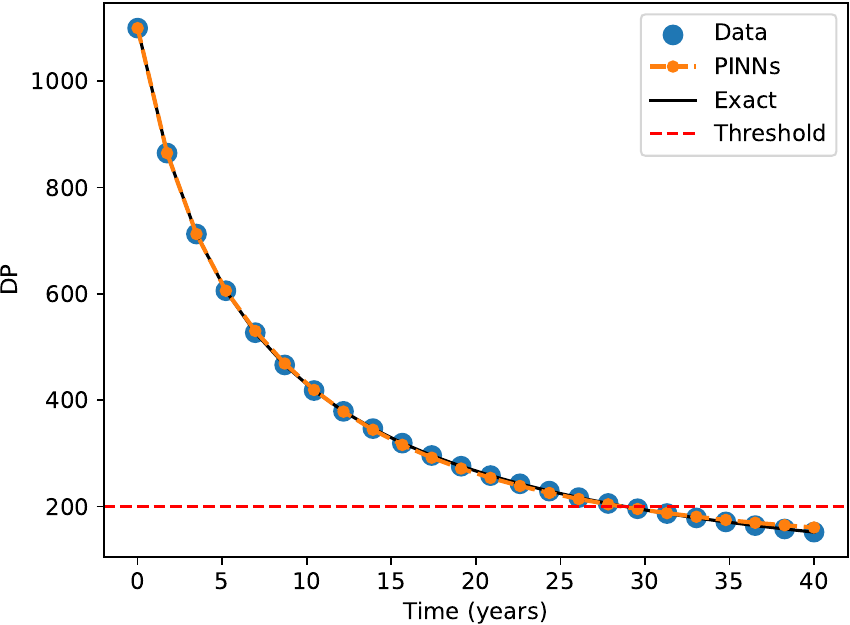}
    \caption{PINNs prediction (orange-dotted line) of Dataset 1 (blue dots) and the corresponding exact solution (black line). The red-dotted line represents the end-of-life of the power transformer.}
    \label{fig:synth_data_pinns}
\end{figure}

\begin{figure}[ht]
    \centering
    \includegraphics[width=1.0\textwidth]{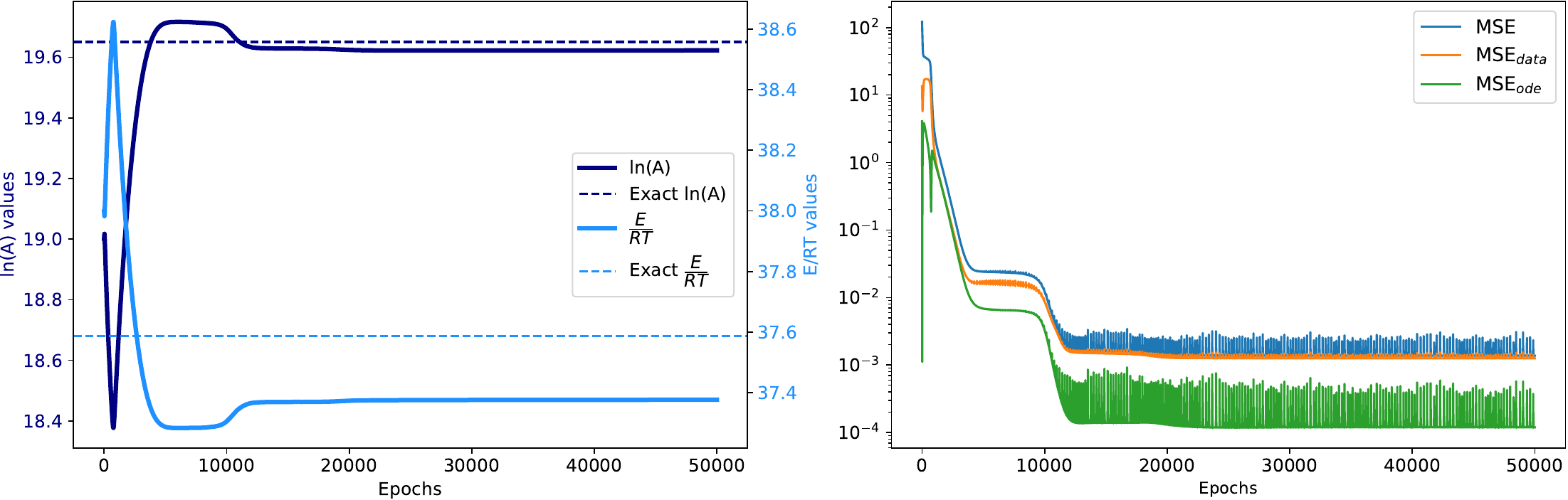}
    \caption{Left: Evolution of the inferred scaled parameters $ln(A)$ (blue solid line) and $\frac{E}{RT}$ (light blue solid line) over the epochs and the corresponding exact values (dotted lines) for Dataset 1. Right: Loss functions over the epochs.}
    \label{fig:params_losses_epochs}
\end{figure}

Fig.~\ref{fig:synth_data_pinns} shows Dataset 1, the blue dots, the exact solution represented with a black line, and the prediction of PINNs with the orange dotted line. The red-dotted line defines the threshold and is set at 200, which we consider the end-of-life of the power transformer. Fig.~\ref{fig:params_losses_epochs} displays a plot of the evolution of the scaled parameters over the epochs on the left, while the loss functions are plotted on the right. The former consists of two solid lines, where the darker blue shows the values of $ln(A)$ and the lighter blue represents $\frac{E}{RT}$, with corresponding dotted lines presenting the exact values with associated colors.

\begin{figure}[ht]
    \centering
    \includegraphics[width=0.65\textwidth]{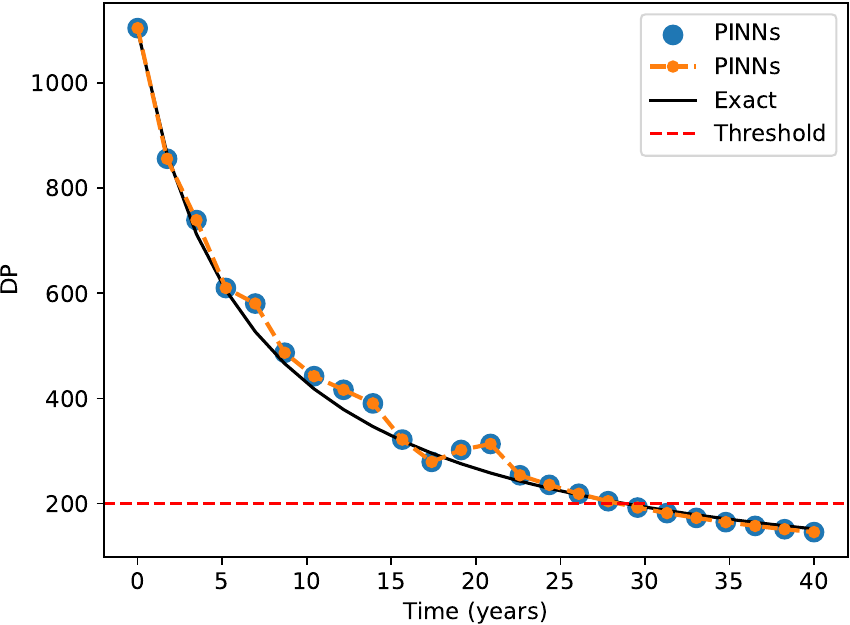}
    \caption{PINNs prediction (orange-dotted line) of Dataset 1 with 10\% noise (blue dots) and the corresponding exact solution (black line). The red-dotted line represents the end-of-life of the power transformer.}
    \label{fig:synth_data_pinns24_10perc}
\end{figure}

\begin{figure}[ht]
    \centering
    \includegraphics[width=1.0\textwidth]{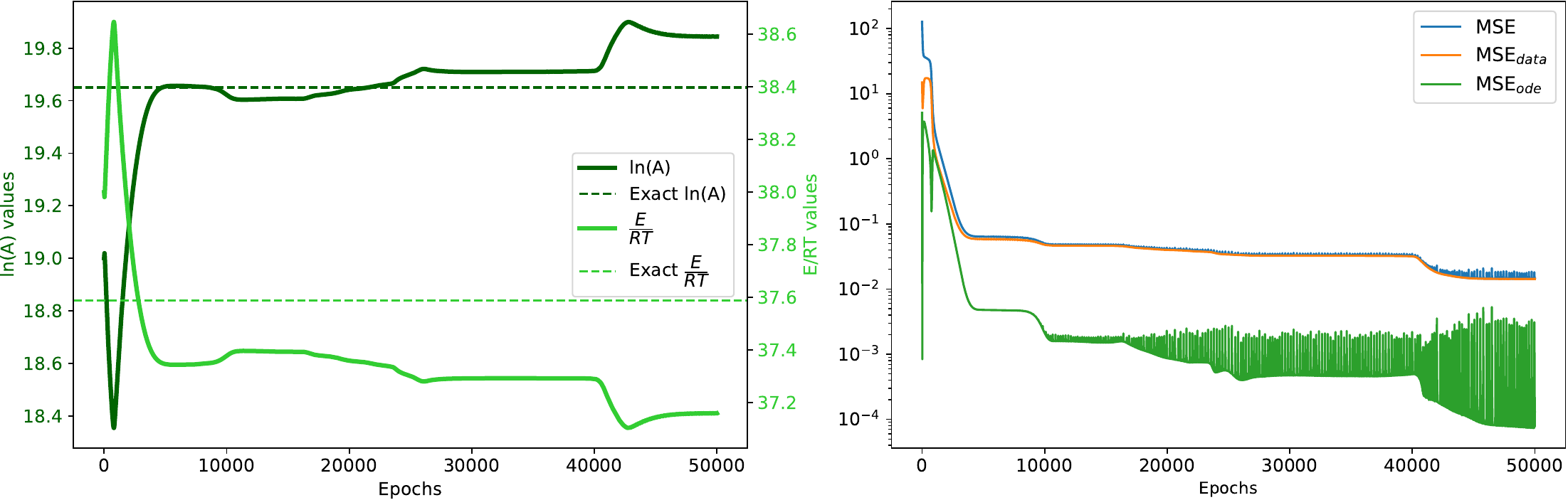}
    \caption{Left: Evolution of the inferred scaled parameters $ln(A)$ (green solid line) and $\frac{E}{RT}$ (light green solid line) over the epochs and the corresponding exact values (dotted lines) for Dataset 1 with 10\% noise. Right: Loss functions over the epochs.}
    \label{fig:params_losses_epochs24_10noise}
\end{figure}

Fig. \ref{fig:synth_data_pinns24_10perc} shows the DP data for Dataset 1 with $10\%$ noise. The blue dots show the noisy data, the black line is the exact solution without the noise, and the orange-dotted line shows the fitting of PINNs to the data. We have noticed that PINNs have started to overfit the noisy data. This can also be seen in the right plot of Fig.~\ref{fig:params_losses_epochs24_10noise}, where the loss functions are displayed. Looking at the left plot of Fig.~\ref{fig:params_losses_epochs24_10noise}, the estimated values of $\frac{E}{RT}$, the lighter green solid line, are quite off from the actual value, represented by the lighter green dotted line. Similarly, for the estimated values of $ln(A)$, in darker green lines. Moreover, both lines jump more around, compared to the ones of the left plot of Fig.~\ref{fig:params_losses_epochs}, which attain convergence just after 10000 epochs.

\begin{figure}[ht]
    \centering
    \includegraphics[width=0.65\textwidth]{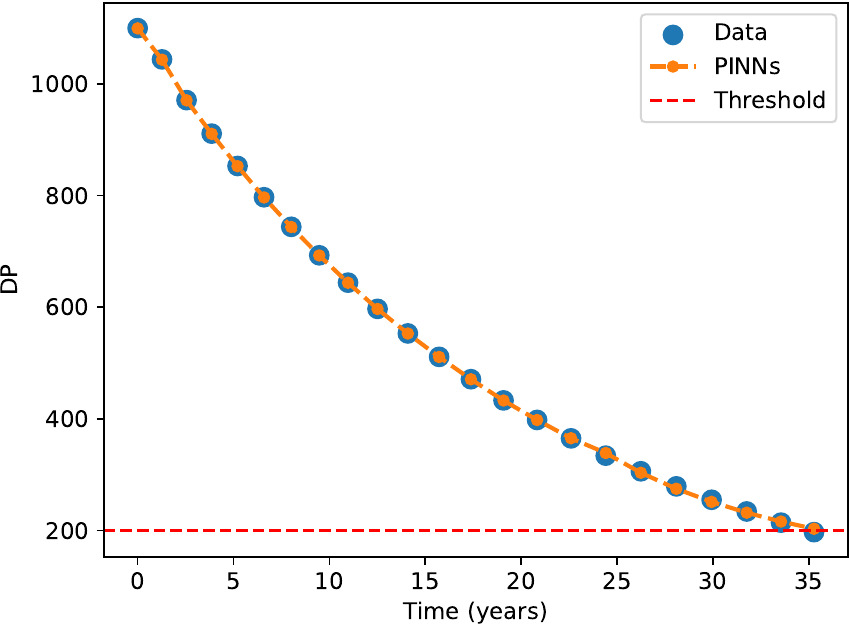}
    \caption{PINNs prediction (orange-dotted line) of Dataset 3 (blue dots). The red-dotted line represents the end-of-life of the power transformer.}
    \label{fig:real_data_pinns}
\end{figure}

\begin{figure}[ht]
    \centering
    \includegraphics[width=1.0\textwidth]{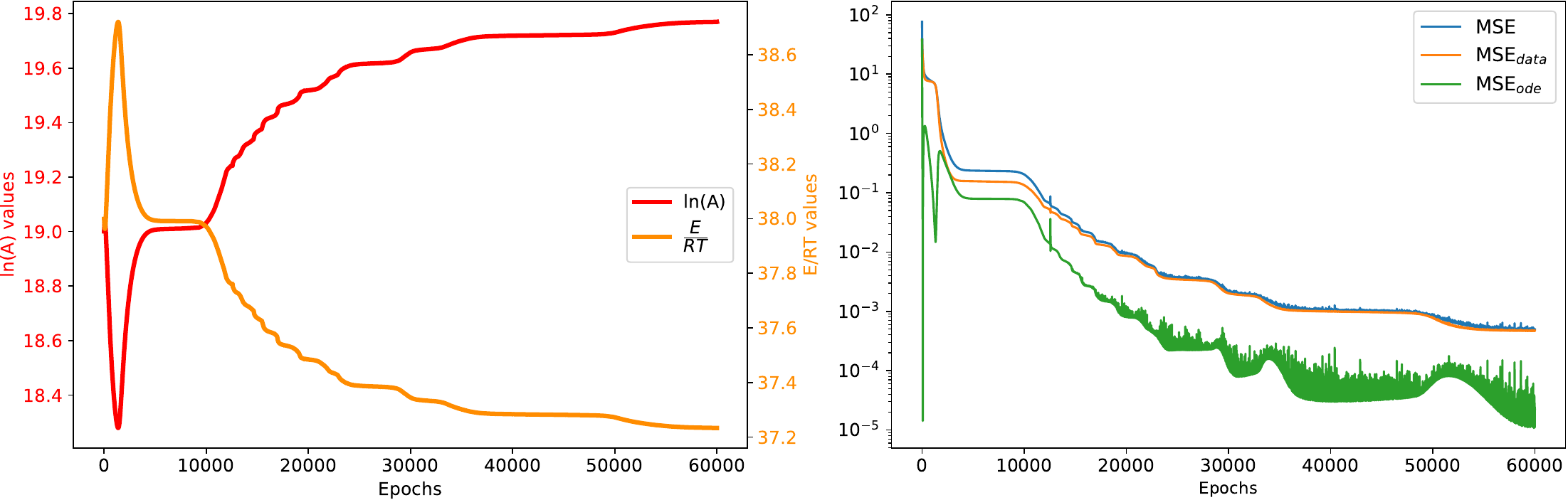}
    \caption{Left: Evolution of the inferred scaled parameters $ln(A)$ (red line) and $\frac{E}{RT}$ (orange line) over the epochs for Dataset 3. Right: Loss functions over the epochs.}
    \label{fig:params_losses_epochs_real}
\end{figure}

Figs.~\ref{fig:real_data_pinns} and~\ref{fig:params_losses_epochs_real} display the estimations obtained for Dataset 3. For this case, we only have the measurements available, but no exact solution can be retrieved as we do not know the values of the parameters $A$ and $E$. From Fig.~\ref{fig:real_data_pinns}, it can be seen that the PINN model fits well with the given data points. The left plot of Fig.~\ref{fig:params_losses_epochs_real} shows the evolution of the scaled unknown parameters, where the red line indicates the value $ln(A)$ while the orange line represents the value $\frac{E}{RT}$. After some increasing steps for the value $ln(A)$ and decreasing steps for the $\frac{E}{RT}$, it seems both values smoothly stabilize around 40000 epochs.

\begin{figure}[ht]
    \centering
    \includegraphics[width=1.0\textwidth]{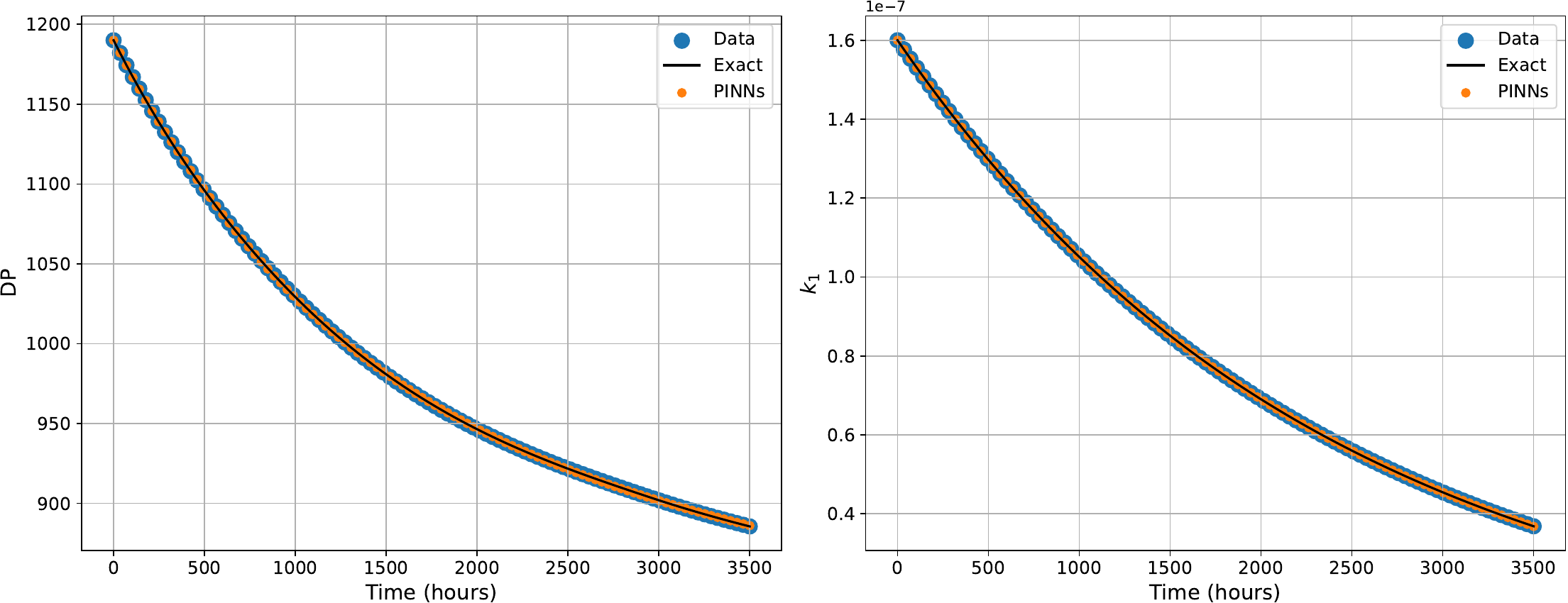}
    \caption{Left: PINNs predictions of DP (orange dots) compared to the data points (blue dots) and the exact solution (black line). Right: PINNs predictions of $k_1$ (orange dots) compared to the data points (blue dots) and the exact solution (black line).}
    \label{fig:system_ident_DPk1}
\end{figure}

\begin{figure}[ht]
    \centering
    \includegraphics[width=1.0\textwidth]{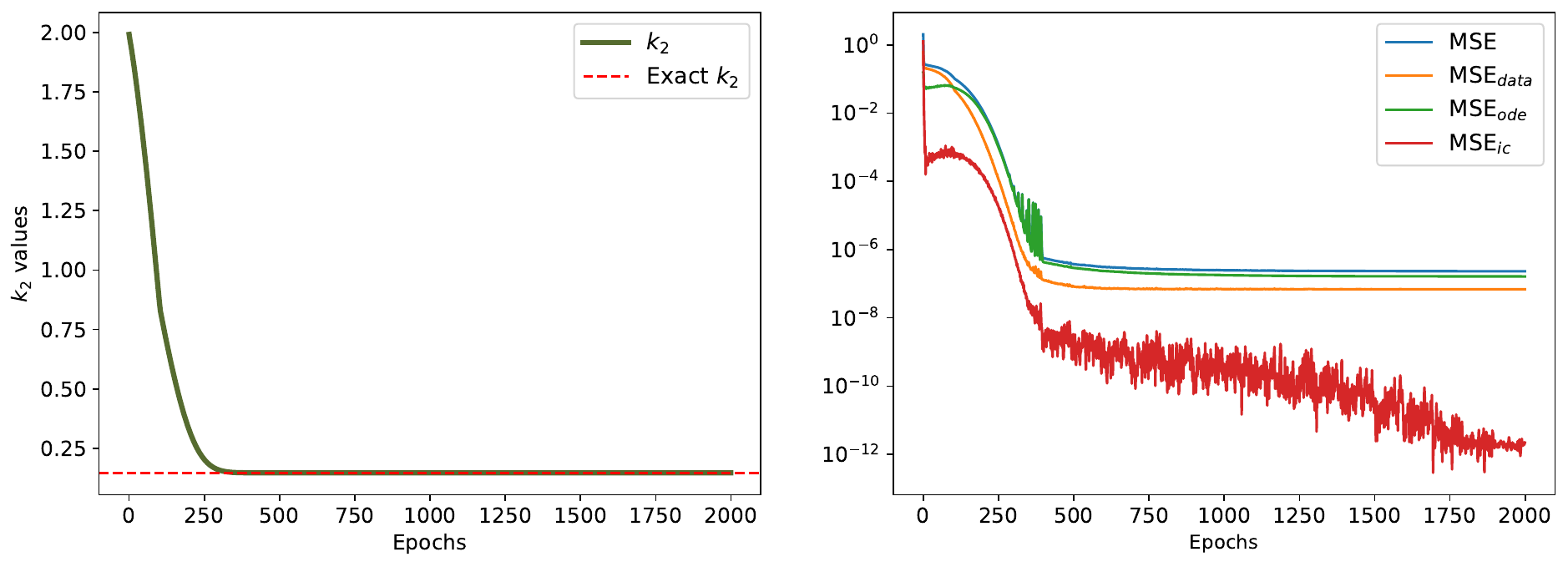}
    \caption{Evolution of the inferred scaled parameter $k_2$ (dark green line) over the epochs, and the corresponding exact scaled value (red-dotted line). Right: Loss functions over the epochs.}
    \label{fig:param_losses_epochs_systemident}
\end{figure}

Now, we focus on the results obtained with PINNs and symbolic regression for discovering one of the parameters and one of the two functions expressing Emsley's system of ODEs. We start by describing the PINNs architecture. The first NN has one neuron for the input layer consisting of the time coordinates and 13 hidden layers with 35 neurons each. The activation function is the hyperbolic tangent. The output layer consists of two neurons approximating the $DP$ and $k_1$ values at the inputted time coordinates. The second NN takes the outputs of the first NN as inputs in addition to the time coordinates. It has 12 hidden layers with 53 neurons, each equipped with hyperbolic tangent activation functions, and outputs the unknown function $h$. The networks are optimized using an Adam optimizer with a learning rate of 1e-3. The model is run for 2000 epochs. In addition to the 1000 simulated data points, 10000 collocation points are generated within the time domain. The initial guess for the scaled unknown value is 2, while the real scaled value is 0.147.
In Fig.~\ref{fig:system_ident_DPk1}, the PINNs predictions for the DP and $k_1$ values are shown with orange dots on the left and right plots, respectively, compared to the training data, the blue dots, and the exact solution, the black line. PINNs predict well the two solutions with a relative $L_2$ error of $2.821\cdot 10^{-4}$ for the DP values and $1.679\cdot 10^{-4}$ for the $k_1$ values. Fig.~\ref{fig:param_losses_epochs_systemident} shows the plot of the progress of the identified scaled parameter $k_2$ on the left, which converges to the actual value relatively fast after 250 epochs. The scaled and unscaled values of $k_2$ are 0.147 and $4.2\cdot 10^{-4}$, while the corresponding identified values are 0.14704 and $4.2012\cdot 10^{-4}$. The percentage error between the estimated value and the exact one is 0.028\%. PINNs estimate very accurately the unknown parameter. In Fig.~\ref{fig:param_losses_epochs_systemident}, on the right, we see the evolution of the loss functions, including the loss function assigned to the initial condition, $\text{MSE}_{ic}$. 

\begin{figure}[ht]
    \centering
    \includegraphics[width=1.0\textwidth]{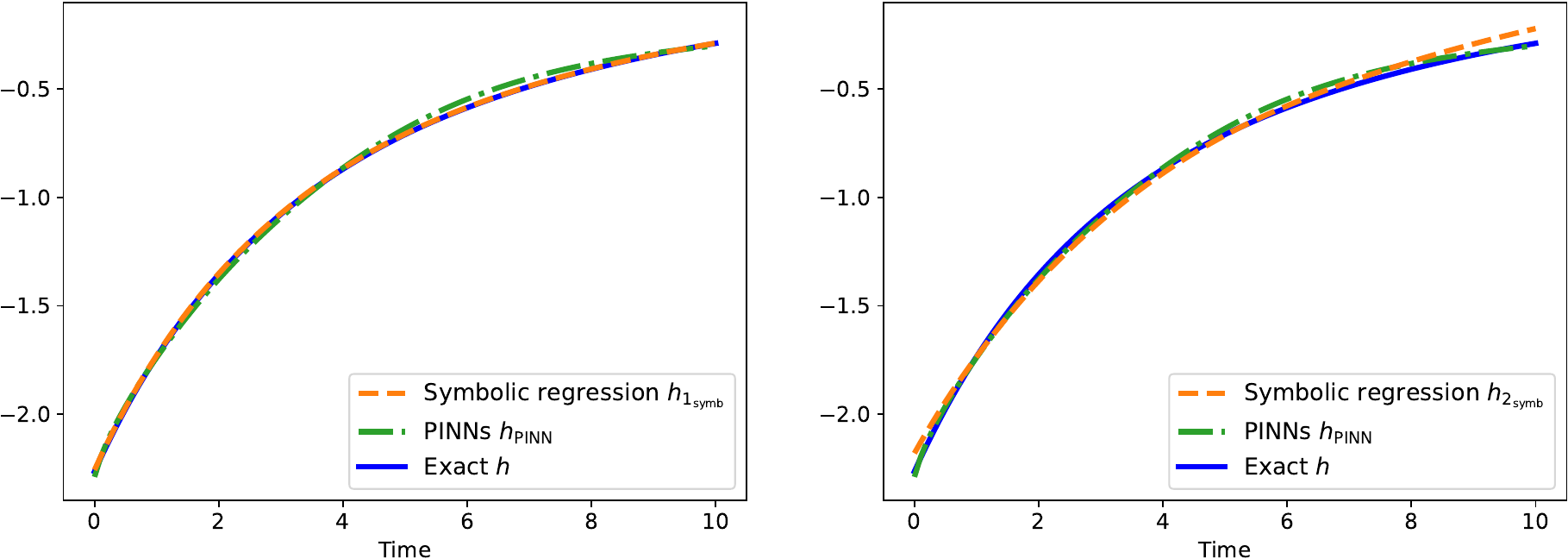}
    \caption{The exact form of equation $h$ (blue line), the values with the inferred result by PINN, $h_{\text{PINN}}$ (green-dotted line), and the identified symbolic regression approximation $h_{\text{symb}}$ (orange-dotted line). Left: Results using only the multiplication. Right: Results using both multiplication, addition, and subtraction.}
    \label{fig:h_exact_pinn_sr}
\end{figure}

The predicted values of $h$ by PINNs, which, for simplicity, we will call $h_{\text{PINN}}$, are plotted in Fig. \ref{fig:h_exact_pinn_sr} with a green dotted line. The relative $L_2$ error between the predictions and the actual functional form of the equation is $2.263\cdot 10^{-2}$. The estimated values for $h_{\text{PINN}}$, DP, and $k_1$ are then fed to the symbolic regression model. We test two cases: one includes only multiplication in the considered binary operations of the mode; the other introduces addition and subtraction. We will define the corresponding symbolic regression approximations of the function $h$ as $h_{1_{\text{symb}}}$ and $h_{2_{\text{symb}}}$. Fig. \ref{fig:h_exact_pinn_sr} displays the plots for the cases considered. On the left, we have the case with only multiplication; on the right, the case with multiplication, addition, and subtraction is considered. The blue lines show the exact form of equation $h$, the green-dotted lines are the PINNs approximation, $h_\text{PINN}$, while the orange-dotted lines present the two symbolic regression approximations, i.e. $h_{1_{\text{symb}}}$ and $h_{2_{\text{symb}}}$. 

\begin{figure}[ht]
    \centering
    \includegraphics[width=0.95\textwidth]{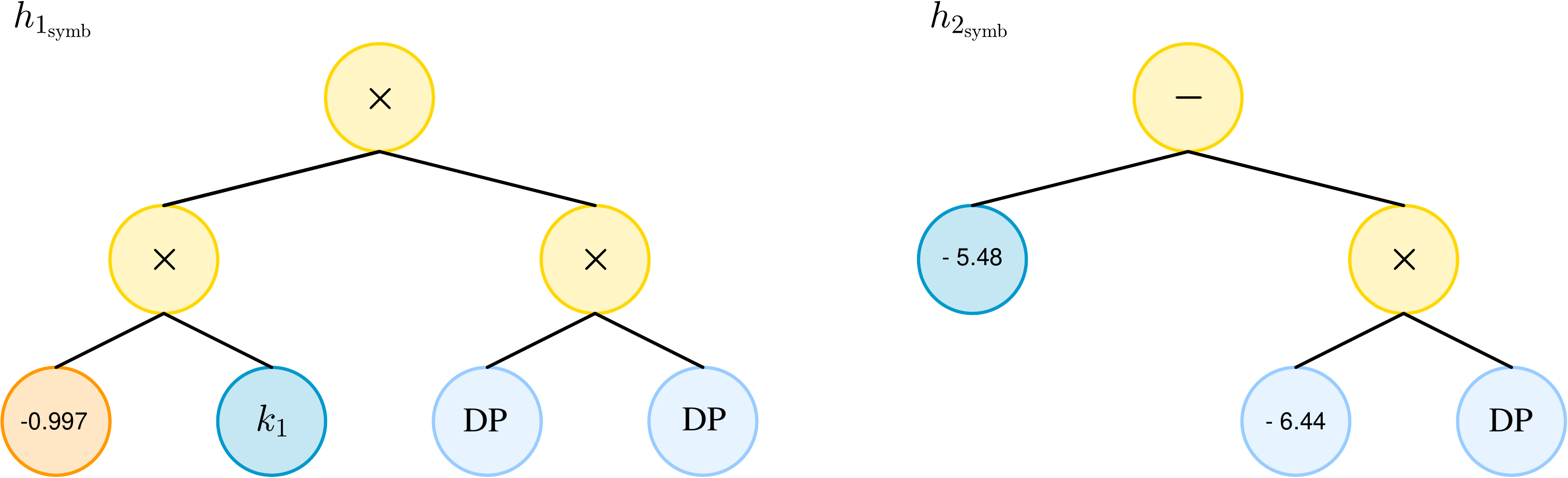}
    \caption{Expression trees to represent the detected functions $h_{1_{\text{symb}}}$ (left) using only multiplication and $h_{2_{\text{symb}}}$ (right) using multiplication, addition and subtraction.}
    \label{fig:expression_tree}
\end{figure}

The inferred mathematical expressions discovered by symbolic regression for the two cases are given by:
\begin{align}
    & h_{1_{\text{symb}}} = -0.997\cdot k_1 \cdot \text{DP}^2, \label{eq:h1_symb} \\
    & h_{2_{\text{symb}}} = 5.48 - 6.44\cdot \text{DP}.
    \label{eq:h2_symb}
\end{align}
Fig.~\ref{fig:expression_tree} displays the corresponding expression trees.
We recall from Eq.~\eqref{eq:emsley_syst} the exact form of the function we would like to recover, which is $h=-k_1\cdot \text{DP}^2$. Therefore, we see that Eq.~\eqref{eq:h1_symb} restores the functional form of the unknown equation very accurately, with a relative $L_2$ error of $3.235\cdot 10^{-3}$. This accurate estimation can also be observed from the left plot of Fig.~\ref{fig:h_exact_pinn_sr}. However, when introducing additional binary operations, such as addition and subtraction, the form of the expression changes to Eq.~\eqref{eq:h2_symb}. These predictions' relative $L_2$ error compared to the exact value is $3.233\cdot 10^{-2}$. The accuracy of the results is still high; however, it has a very different expression that does not even consider the values of the solution for $k_1$, present in the original function.

\section{Discussion}
\label{sec:discussion}
In this section, we discuss the obtained results and delineate the steps more carefully to discover partially known ODEs. Moreover, we also state the advantages and drawbacks of using the methods utilized in this paper. We start by considering the discovery of unknown parameters of the Ekenstam equation. We show the possibility of employing a model like PINNs to infer unknown parameters of the equation, particularly in this case, the pre-exponential factor and the activation energy of the Arrhenius equation. First, the considered parameters are very large; therefore, it is necessary to have some scaling of the equation to help mitigate the problem of PINNs handling such values. For our problem, we end up using the natural logarithm of the equation, which needs careful handling as we need to ensure that the values are not negative. For example, the choice of activation energy of the NN is fundamental since a function like hyperbolic tangent, which is the most adopted in PINN models, does not work as it squeezes the values in an interval between -1 and 1. Therefore, a choice of sigmoid function is preferred as it takes positive values between 0 and 1. The data type used is of major interest to understand the values and their relationship. We analyzed two synthetic datasets with added noises and one measurement dataset. We can see that PINNs always range within reasonable predictions of unknown parameters in all cases. However, when noise is added, the model lacks robustness and starts to overfit. A way to fix this issue is to use Bayesian Physics-Informed Neural Networks (BPINNs)~\cite{yang2021b}, which can quantify aleatoric uncertainty within the Bayesian framework while predicting the solution from large, noisy data. BPINNs have been proven to solve accurately forward and inverse problems for real-world nonlinear dynamical systems~\cite{linka2022bayesian}; therefore, it could be useful in the DP analysis to quantify the uncertainty from a transformer. We could also notice that more data does not necessarily mean more accurate results. From the error analysis of the results of the two synthetic data with 24 and 48 data points, we could see that the smaller dataset slightly performs better. This is one of the main advantages of PINNs, which do not require much data to train, like classical ANNs, because they exploit the knowledge given by the provided physical system. In the case of modeling degradation and aging of paper inside transformers, cellulose aging data is scarce. Therefore, being able to use a model like PINNs can be favorable. Given any DP measurements, our model can identify the corresponding values of the pre-exponential factor $A$ and the activation energy $E$ of the Arrhenius equation. Note that we already have an idea from previous studies of the value range of these parameters. To make sure to lie within the range, we can either control the initialization of the parameters, as we do in this work or set up some constraints given some minimum and maximum values within the network. If this is ignored, the parameters estimated can be quite far from the actual ones.  

Moving to discover the unknown function and parameter of Emsley's system of ODEs part, we can claim that the model already performs well. In particular, the unknown parameter $k_2$ is accurately identified with a low percentage error compared to the ground truth. However, there are some limitations that we can point out. First, the hyperparameter exploration needs to be very detailed, which can be time-consuming, and, in this case, knowing the original functional form of the equation is quite biased. The problem arises because our unknown $h(DP,k_1,t)$ has some parabolic-looking curve shape that many mathematical expressions could approximate. Therefore, if we do not have a very accurate approximation of the unknown function already using PINNs, we will end up with several different expressions that could be very close to the actual one. We set up 13 hidden layers with 35 neurons each for the first network, while we have 12 hidden layers with 53 neurons for the second one. These architectures are unusual but best-performing for this problem within the range of tested architectures.
Interestingly, the choice of symbolic regression also plays a fundamental role in getting the desired results. After feeding the symbolic regression with DP and $k_1$ values estimated by PINNs along with the inferred unknown function used as a target, we must choose the type of operations the model needs to explore. If we choose only multiplication, we obtain almost the exact function form. However, the symbolic regression model identifies a different function when we propose several operations from which to search. This is again related to the fact that the function can be approximated in various ways. Therefore, an accurate analysis and model setup is crucial to obtain accurate results. The results are case-centered, but a more generalizable model could help further investigation using different initial values of DP and $k_1$ and values of $k_2$. Despite getting the wrong equation, this can lead to more precise identification of the equations that describe the degradation and aging mechanisms in power transformers. Given the proper measurements, it will be possible to infer some novel equations that could be tested afterward with further measurements and experiments. 

Another possible method not investigated in this paper that could be implemented for identifying the unknown system is in the case of a partially known equation. We know the exact form of one part of the ODE, but the relationship is not entirely captured, and the form of the remaining function needs to be inferred. For example, we could assume there exists some additional function that we could try to approximate of the Ekenstam equation. We could verify if the already used form captures all the system behaviors or if there could be additional parts that could improve the model's accuracy.

These machine learning models that embed the physical knowledge of equations do not just solve the problem of finding the unknown parameters and functions alone. They must be accurately applied when measurements from in-operation transformers and lab experiments are available for specific estimations. Then, they need to be tested with further measurements that can improve the estimations. It is a continuous learning process from the PINNs and symbolic regression models that must be validated with field data.

\section{Conclusion}
\label{sec:conclusion}

In this paper, we analyze two different ways to discover partially known ODEs. First, we identify the unknown parameters of the Ekenstam equation, where the Arrhenius equation defines its kinetic rate using PINNs. The inferred parameters lie within the range of values reported in the literature. When adding noise, the results have a larger variance, and the model does not perform as well. We tested one case with measurements, and the results agree with the literature and other findings. Then, we discover an unknown function and parameter of Emsley's system of ODEs. The model can recover the unknown parameter. Two cases with symbolic regression to find the functional form are tested, showing that the given equation has several ways to be approximated. More generalizable models need to be investigated. 

In general, the resulting models allow the discovery of unknown parameters from the limited data set. For the discovery of the function, the results lead to other equation forms requiring additional experimental tests and field data collection to validate.
The findings exposed in the paper should serve as possible guidance in modeling cellulose aging in power transformers and, more generically, system identification.

\section*{Acknowledgments}
This work is part of the project "Using Physics-Informed Machine Learning for reusing power system components" funded by the "Circular and bio-based economy" program from Vinnova (The Swedish Innovation Agency).